\pdfoutput=1

\documentclass[11pt]{article}

\usepackage{acl}

\usepackage{times}
\usepackage{latexsym}

\usepackage[T1]{fontenc}

\usepackage[utf8]{inputenc}

\usepackage{microtype}

\usepackage{inconsolata}
\usepackage{amsmath}
\usepackage{bbding}
\usepackage{booktabs}
\usepackage{comment}
\usepackage{graphicx}
\usepackage{microtype}
\usepackage{mathtools}
\usepackage{multicol}
\usepackage{multirow}
\usepackage{pifont}
\usepackage{subcaption}
\usepackage{scrextend}
\usepackage{setspace}
\usepackage{xcolor}
\usepackage{xspace}

\usepackage{array}
\newcolumntype{L}{>{\centering\arraybackslash}m{3cm}}
\newcommand{\texto}[1]{{\fontfamily{bch}\selectfont{{#1}}}}
\newcommand{\neurologic}{\texto{NeuroLogic}\xspace}
\newcommand{\score}{$S$\xspace}
\newcommand{\neurocomps}{\texto{NeuroComparatives}\xspace}
\newcommand{\neurocompsshort}{\texto{NCs}\xspace}
\newcommand{\ncsmall}{\texto{NC-S}\xspace}
\newcommand{\neurocompsl}{\texto{NC-L}\xspace}
\newcommand{\neurocompsxl}{\texto{NC-XL}\xspace}

\newcommand{\cmark}{\textcolor{green}{\ding{51}}}%
\newcommand{\xmark}{\textcolor{red}{\ding{55}}}%

\newcommand{\draftonly}[1]{#1}
\renewcommand{\draftonly}[1]{}

\newcommand{\draftcomment}[1]{\draftonly{#1}}
\newcommand{\swabha}[1]{\draftcomment{\textcolor{orange}{\small [#1]$_{{SS}}$}}}
\newcommand{\phillip}[1]{\draftcomment{\textcolor{blue}{\small [#1]$_{{PH}}$}}}

\newcommand{\remove}[1]{}

\title{
\neurocomps: Neuro-Symbolic Distillation of Comparative Knowledge}

\newcommand{\aspace}{\hspace{1em}}
\newcommand{\uw}{$^{\heartsuit}$}
\newcommand{\intel}{$^{\diamondsuit}$}
\newcommand{\duke}{$^{\dagger}$}
\newcommand{\aiTwo}{$^{\clubsuit}$}
\newcommand{\usc}{$^{\spadesuit}$}
\newcommand{\equal}{$^{\ast}$}

\author{
    Phillip Howard\intel\equal\aspace 
    Junlin Wang\duke\equal\aspace 
    Vasudev Lal\intel \aspace 
    Gadi Singer\intel\aspace \\
    \textbf{Yejin Choi}\uw\aiTwo \aspace 
    \textbf{Swabha Swayamdipta}\aiTwo\usc\aspace \\
    \intel Intel Labs \aspace \aiTwo Allen Institute for AI \aspace \usc University of Southern California\\
    \duke Duke University \\
    \uw Paul G.\ Allen School of Computer Science \& Engineering, University of Washington \\
    \texttt{phillip.r.howard@intel.com}
}

\newcommand\blfootnote[1]{%
  \begingroup
  \renewcommand\thefootnote{}\footnote{#1}%
  \addtocounter{footnote}{-1}%
  \endgroup
}

\begin{document}
\maketitle
\blfootnote{\equal Equal contribution}

\begin{abstract}

\emph{Comparative knowledge} (e.g., steel is stronger and heavier than styrofoam) is an essential component of our world knowledge, yet understudied in prior literature. 
In this paper, we harvest the dramatic improvements in knowledge capabilities of language models into a large-scale comparative knowledge base. 
While the ease of acquisition of such comparative knowledge is much higher from extreme-scale models like GPT-4, compared to their considerably smaller and weaker counterparts such as GPT-2, not even the most powerful models are exempt from making errors.
We thus ask: to what extent are models at different scales able to generate valid and diverse comparative knowledge?

We introduce \neurocomps, a novel framework for comparative knowledge distillation overgenerated\footnote{We use “overgenerate” throughout this work to indicate that we generate more knowledge from the language model than we intend to keep after rigorous filtering and selection.} from language models such as GPT-variants and LLaMA, 
followed by stringent filtering of the generated knowledge.
Our framework acquires comparative knowledge between everyday objects, producing a corpus of up to 8.8M comparisons over 1.74M entity pairs---10X larger and 30\% more diverse than existing resources. 
Moreover, human evaluations show that \neurocomps outperform existing resources in terms of validity (up to 32\% absolute improvement).
Our acquired \neurocomps leads to performance improvements on five downstream tasks.
We find that neuro-symbolic manipulation of smaller models offers complementary benefits to the currently dominant practice of prompting extreme-scale language models for knowledge distillation.
\end{abstract}

\section{Introduction}
\label{sec:intro}
\begin{figure*}[ht!]
\centering
  \includegraphics[width=1\textwidth]{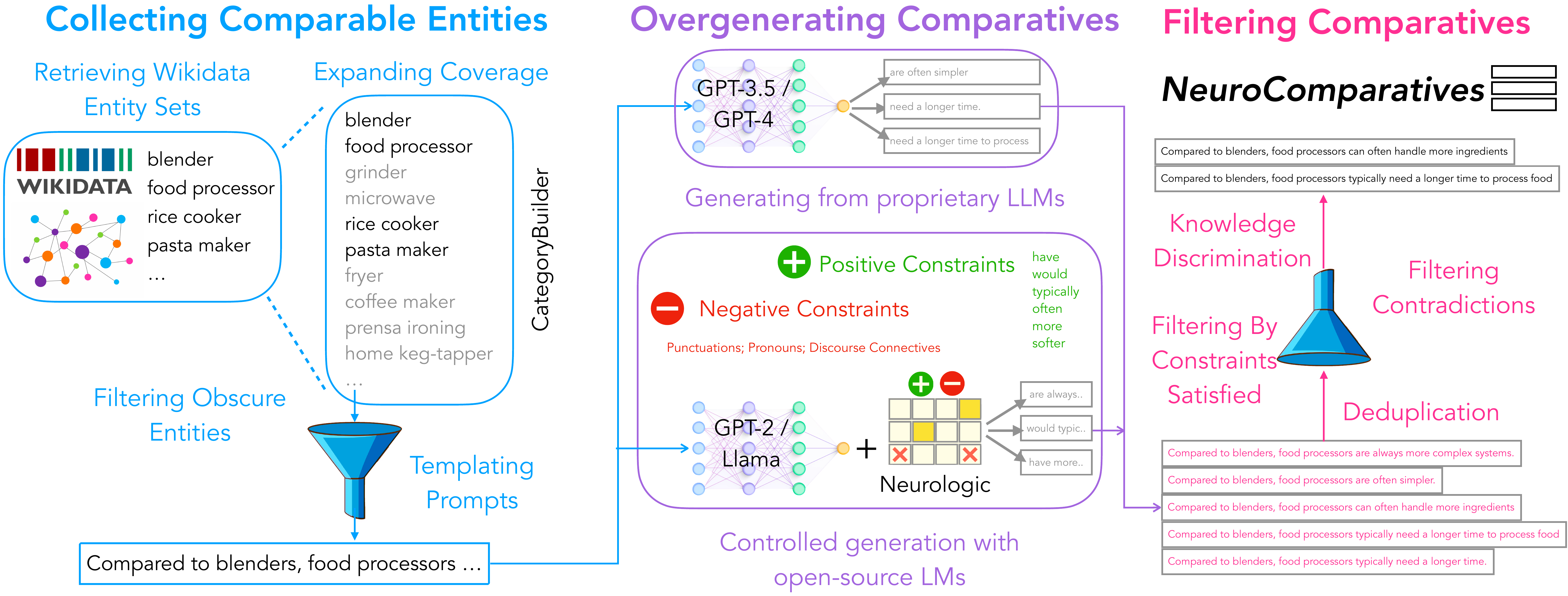}
  \caption{Our neuro-symbolic framework to distill \textbf{\neurocomps}. 
  (1) We seed entity pairs for comparison from Wikidata, and expand the set with CategoryBuilder to construct templated prompts for a language model.
  (2) Next, we use these prompts to overgenerate comparatives from different language models to ensure our generations contain valid comparisons between a given pair of entities. 
  (3) Finally, we discard contradictory and otherwise lower quality generations via various clustering and filtering techniques.
  Our resultant corpus \neurocomps contains 8.8 million comparisons over 1.74 million entity pairs. \remove{, comparable in quality to GPT-3 generated comparatives.}
  }
  \label{fig:overview}
\end{figure*}

In their book \emph{``Surfaces and Essences''} on \emph{concepts} and \emph{analogies}, \citet{hofstader2013surfaces} elucidate how concept learning requires \emph{comparing} a pair of concepts, and parsing out their similarities and dissimilarities. 
Indeed, comparative knowledge is an essential component of our world knowledge \cite{ilievski2021dimensions,davis2023benchmarks}, underpinning some of the classical commonsense reasoning problems. 
For example, the problem \emph{``The large ball crashed right through the table because it was made of [steel/styrofoam]. What was made of [steel/styrofoam]?''} in Winograd Schema Challenge \cite{levesque2011winograd} requires comparing the relative strength between steel and styrofoam.  
Yet, compared to \emph{general} knowledge acquisition, there has been relatively little research focus on \emph{comparative} knowledge acquisition, possibly due to the longstanding challenges of high-quality knowledge acquisition itself, let alone comparative knowledge.
The few resources for comparative knowledge, all derived from web mining, however limited in size and diversity (\S\ref{sec:related}), have nonetheless been useful for challenging multimodal reasoning tasks \cite{wang2018fvqa}, highlighting the value of comparative knowledge.

In this paper, we draw inspirations from such literature 
about concept learning and inquire two related questions on \emph{comparative knowledge}:
(1) 
how well do models at different scales fare at the task of producing large-scale, high-quality comparative knowledge about a broad range of concepts? and
(2) what are the implications for downstream tasks?
Compared to prior resources of commonsense knowledge acquired either via crowdsourcing \cite{speer2017conceptnet,sap2019atomic} or via information extraction (e.g., WebChild \cite{tandon2017webchild} and ASER \cite{zhang2020aser}), our attempt to (re-)focus on the task of comparative knowledge acquisition takes the perspective of \emph{``language models as knowledge bases''} \cite{west2021symbolic, alkhamissi2022review}, motivated by the dramatic improvements in the capabilities of extreme-scale neural language models. 

We build on such research, but in addition to few-shot inference with extreme-scale model APIs such as GPT-4, also rely on customized inference with smaller language models \cite{see-etal-2019-makes,sheng-etal-2020-towards,liu-etal-2021-dexperts}. 
We ask a seemingly implausible question: can  considerably smaller and weaker language models such as GPT-2 \cite{Radford2019LanguageMA}, complement the capabilities of their large-scale counterparts in the acquisition of \emph{comparative knowledge} between a pair of concepts?
To this end, we follow an overgenerate-and-filter mechanism \cite{langkilde-knight-1998-generation-exploits,walker-etal-2001-spot} to create a large-scale, high-quality resource: \textbf{\neurocomps}, a corpus with up to 8.8 million comparisons over 1.74 million pairs of entities. 
Our framework is illustrated in \autoref{fig:overview}.

Compared to the only other large-scale commonsense KG containing comparative knowledge \cite[WebChild]{tandon2017webchild}, \neurocomps is up to 10x larger, 30\% more diverse, and has a 19\% higher human acceptance rate. 
Additionally, we show that a knowledge discriminator model can further improve the the human acceptance rate of our knowledge to 90\%, representing a 32\% absolute gain compared to WebChild while still being over 2X larger in scale. 
Our analyses also show that \neurocomps are, on aggregate, more diverse than WebChild comparatives and more effective on three different downstream benchmarks.
Overall, our findings motivate customizable neuro-symbolic manipulation of smaller scale  models as a cost-effective complement to the dominant practice of performing simple inferences under extreme-scale yet closed language models. 
We make our code\footnote{\url{https://github.com/IntelLabs/multimodal_cognitive_ai/tree/main/NeuroComparatives}} and dataset\footnote{\url{https://huggingface.co/datasets/Intel/NeuroComparatives}} publicly available.

\section{Distilling \neurocomps}
\label{sec:method}

Our framework for distilling comparatives from an autoregressive LM comprises three stages, illustrated in Figure~\ref{fig:overview}.
First, we collect comparable entities to construct prompts for eliciting comparative knowledge statements
(\S\ref{sec:creating-prompts}).
Next, we employ LMs to overgenerate (potential) comparatives for every pair of selected entities (\S\ref{sec:overgeneration}).
Finally, we filter the generations (\S\ref{sec:filtering}) to obtain a large-scale, high-quality collection of comparative statements, which we call \textbf{\neurocomps (\neurocompsshort)} (\S\ref{sec:neurocomparatives}). 

\subsection{Collecting Comparable Entities}
\label{sec:creating-prompts}
One unique challenge in probing language models for knowledge acquisition, as opposed to extracting pre-existing knowledge from web text, is knowing \emph{exactly what} to probe LMs for, i.e., concept pairs for comparison. 
For practical applications, the comparatives are more likely to be useful when they are about entities sharing some common properties, e.g., ``red wine'' and ``white wine'' (Fig.~\ref{fig:ordered-constraint-satisfaction}), rather than unrelated ones e.g., ``cucumber'' and ``car''. 
Identifying a vast array of diverse yet relevant concepts for comparison is complex; thus, we developed a systematic process below: retrieval (\S\ref{sec:initial-entities}), expansion (\S\ref{sec:expanded-entities}), and filtering (\S\ref{sec:filtered-entities}).


\begin{figure}[t!]
    \centering
    \includegraphics[width=0.97\columnwidth]{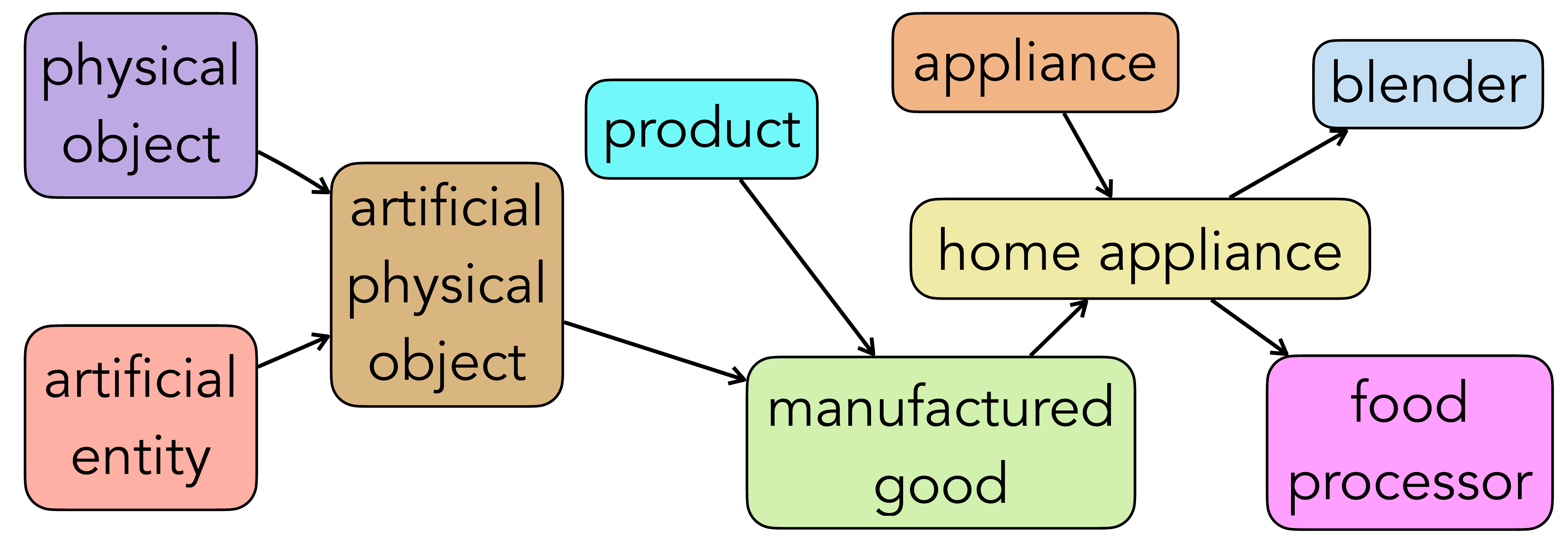}
    \caption{Wikidata hierarchical class structure for retrieved entities \lq blender' and \lq food processor'. 
    }
    \label{fig:wikidata-structure}
\end{figure}

\subsubsection{Retrieving Seed Entity Sets}
\label{sec:initial-entities}

We start our entity collection using two broad Wikidata~\cite{vrandevcic2014wikidata} classes,  as our seed classes: \emph{physical object} 
and \emph{artificial physical object}.  
Each seed class contains entities and subclasses, which themselves may contain additional entities. 
Figure~\ref{fig:wikidata-structure} illustrates an example Wikidata class structure for  ``blender'' and ``food processor'', with ``physical object'' as the root. 
Using a breadth-first traversal of  Wikidata, we retrieve all classes up to two levels\footnote{
We use a maximum search depth of two based on the observation that descending lower in Wikidata results in entities that are too specific or obscure for generating comparatives.
} below the root class.
Overall, we retrieved 1.5K classes with 23K entities from Wikidata. 
While Wikidata provides a good starting point, we find that many of its classes are incomplete, a common challenge with any taxonomic resource.  
Thus, we next expand our entity sets to increase the coverage.

\subsubsection{Expanding the Coverage of Entity Sets}
\label{sec:expanded-entities}

We expand our entity collection using CategoryBuilder~\cite{mahabal2018robust}, a system for lexical entity set expansion. 
We append each retrieved entity set from Wikidata with the top $n = 100$ related terms identified by Category Builder using the hyperparameter $\rho = 3.0$ for context weighting. 
This results in a total of 40K entities corresponding to 1.5K Wikidata classes.
Noting the presence of obscure entities, e.g., ``home keg tapper'' and ``prensa ironing'' in the ``home appliance'' class (Fig.~\ref{fig:overview}), we then moved to eliminate these obscure entities.

\subsubsection{Filtering Obscure Candidate Entities} 
\label{sec:filtered-entities}

Obscure entities would occur infrequently, thus we discard entities which occurred less than $n=100$ times in the language model's training corpus.\footnote{The only corpus we have access to is the open-source counterpart of GPT-2's training corpus, 
{OpenWebText}.}
We additionally discard all classes with less than 2 entities after this filtering step. These filtering steps are applied twice: first on the original Wikidata entities, and then again after we expand the entity sets with Category Builder.
This results in 568 classes with a total of 15,476 entities. 

\subsubsection{Templating Comparative Prompts} 
\label{sec:prompt-and-constraints}

We generate comparison candidates by pairing entities within each class.
For each such pair, $(\texttt{entity}_{1},$ $\texttt{entity}_{2})$, we use the following template\footnote{We experimented with other templates but found that this one was most consistent at generating valid comparisons} to form the prompt for generation:
\vspace{-.25cm}
\begin{equation}
    \textrm{Compared to $\texttt{entity}_{1}$, $\texttt{entity}_{2}$} \ldots 
\end{equation}
\vspace{-7mm}

\noindent
As a final step, we further filter 30\% of the created prompts using GPT-2 XL perplexity to remove potentially disfluent or nonsensical prompts.
This results in a total of 1,741,962 prompts. 

\subsection{Overgenerating Comparatives}
\label{sec:overgeneration}

Since there's no supervision available, we follow an inference-only process for generating the initial set of comparative statements.
We employ two approaches: constrained decoding with open-source LMs (\S\ref{sec:overgenerating-gpt2}) and few-shot prompting of proprietary LLMs (\S\ref{sec:overgenerating-gpt4}).
We hypothesize that both approaches may offer complementary benefits.

\subsubsection{Generation with open-source LMs}
\label{sec:overgenerating-gpt2}

We experiment with GPT-2 XL and Llama-2 7B, two open-source models where inference can be customized to generate comparatives.
We use a customized controlled decoding algorithm, \neurologic \cite{lu2021neurologic} to guide the generation using the prompts constructed above (\S\ref{sec:creating-prompts}).

\paragraph{Formulating the Constraint Sets}
\label{sec:constraints}

We classify our constraints into three types: \emph{positive}, \emph{negative}, and \emph{comparative adjectives}.
Positive constraints ensure tokens appear in the output; we include auxiliary verbs (e.g., `have', `are', `would') and adverbs of frequency (e.g. `typically', `often')
(Appendix~\ref{app:generation-details} for details).
Negative constraints ensure tokens do not appear in the output to reduce hallucinations; we include certain punctuation characters, pronouns, discourse connectives, and relative clauses(Table~\ref{tab:negative-constraints-full} in Appendix~\ref{app:generation-details} for details).

\paragraph{Constrained decoding with \neurologic}
\label{sec:neurologic}
We adapted \neurologic constrained decoding for fluent text generation under specific lexical constraints (\neurologic accepts \textit{clauses} $\{C_{i} \mid i \in 1, \cdots m\}$ as constraints; see details in App.~\ref{app:neurologic}).

\begin{table*}[ht!]
\resizebox{\textwidth}{!}{
\begin{tabular}{lrr} 
\toprule
\textbf{Prompt} & \textbf{WebChild Assertions} & \textbf{Completions in \neurocomps (Ours)}  \\
\midrule
{Compared to helicopters, planes $\ldots$} & $\ldots$ were cooler \cmark \cmark \xmark & $\ldots$ are more stable in flight \cmark \cmark \cmark \\
& $\ldots$ are noisier \cmark \cmark \xmark & $\ldots$ typically have higher operating costs \cmark \cmark \xmark \\
& $\ldots$ are better \cmark \cmark \xmark & $\ldots$ can often carry more cargo \cmark \cmark \cmark\\
\midrule
Compared to floppy disks, hard drives $\ldots$ & $\ldots$ are better \cmark \cmark \cmark & $\ldots$ are generally considered more reliable \cmark \cmark \cmark \\
\midrule
{Compared to cars, motorcycles $\ldots$} & $\ldots$ are cheaper \cmark \xmark \cmark & $\ldots$ generally have fewer moving parts \cmark \cmark \cmark \\
& $\ldots$ are smaller \cmark \cmark \xmark & $\ldots$ generally have lower fuel consumption \cmark \cmark \cmark \\
& $\ldots$ are cooler \xmark \xmark \xmark & $\ldots$ tend to have shorter range \cmark \cmark \cmark \\
\midrule
Compared to blenders, food processors $\ldots$ & $\ldots$ are larger \cmark \cmark \cmark & $\ldots$ can often be more expensive \cmark \cmark \cmark \\
& $\ldots$ work better \cmark \cmark \cmark & $\ldots$ can often handle more ingredients  \cmark \cmark \cmark \\
\bottomrule
\end{tabular}
}
\caption{Generations from \neurocomps and WebChild assertions for the same entity pair. 
Each example was annotated by three human workers: \cmark indicates acceptance and \xmark \ rejection. 
In contrast to WebChild assertions, \neurocomps can be more specific to the entity pairs under consideration, diverse and less subjective.
 }
\label{tab:qualitative}
\end{table*}

To encourage diversity in the generated comparatives, the generator must use different comparative adjectives, without explicit enumeration.
Hence, we dynamically promote top-$k$ comparative adjectives with the highest probabilities at each decoding time step (we use $k$=5). 
We customize this as a special type of positive constraint (\S\ref{sec:prompt-and-constraints}).

We additionally modify \neurologic to handle ordered constraint satisfaction, for fine-grained control.
For each clause $C_{i}$, we assign one or more order indices $m_{i} \in \{1, ..., m\}$ which correspond to the positional order in which clause $C_{i}$ can appear in the generation. Specifying more than one order index allows a clause to appear in multiple different positions. 
Ordered constraint satisfaction provides more fine-grained control for generating valid comparatives, as illustrated in Figure~\ref{fig:ordered-constraint-satisfaction}.

\subsubsection{Generation with Proprietary LLMs}
\label{sec:overgenerating-gpt4}

Many recent LLMs are proprietary, disabling our ability to customize the decoding process due inability to modify the decoding process. 
Hence, we leverage their in-context learning abilities to overgenerate completions for each entity pair. 
Specifically, we use six in-context examples of comparatives followed by our templated comparative prompt (details in App.~\ref{app-gpt3}). 

\subsubsection{Scale of Overgeneration}

Given its open-source availability and relatively low computational cost, we overgenerate the most comparatives using GPT-2-XL as our base LM.
In total, we perform 30 passes of \neurologic with GPT-2 over the 1.74 million entity pairs from \S\ref{sec:prompt-and-constraints}, where each iteration uses a different combination of the positive constraints, while adhering to the same negative and comparative adjective constraints.
Each pass produces 10 generations, resulting in 300 candidate comparatives for each entity pair. This process produces a total of 522 million overgenerations across the 1.74 million entity pairs.
We similarly use Llama-2-7b with \neurologic to generate comparatives for 50K entity pairs, which produces 15 million overgenerations.

In addition, we utilize three propriety LLMs (InstructGPT, ChatGPT, and GPT-4) to overgenerate comparatives for a smaller set of 2.3k entity pairs due to the much higher inference cost. For each entity pair, we return 128 completions, which is the maximum allowed via the inference API. This produces a total of 300k overgenerated comparatives.

\subsection{Filtering Overgenerated Comparatives}
\label{sec:filtering}
Despite using a combination of effective language generation, we achieve quality control through aggressive filtering of the overgenerated comparatives.
This last filtering step consists of deduplication (\S\ref{sec:deduplication}), filtration by constraint satisfaction (\S\ref{sec:filter-constraints}), filtration of contradictory knowledge (\S\ref{sec:filter-contradictions}), and additional filtering via a knowledge discriminator model (\S\ref{sec:knowledge-discriminator}). 

\subsubsection{Deduplication}
\label{sec:deduplication}

To address LM's tendency to generate redundant comparisons, we deduplicate our generations.
We use agglomerative clustering of all generated comparatives using the inner product of their sentence T5 embeddings \cite{Ni2021SentenceT5SS} as the distance. 
For each cluster, we retain only the generation with the best decoding score, \score\footnote{For proprietary LLMs, we use length-penalized perplexity from GPT-2 XL in lieu of the decoding score, \score.} (App.~\ref{app:neurologic}).
Approximately 17\% of the original generations remain. 

\subsubsection{Filtration by Constraint Satisfaction}
\label{sec:filter-constraints}

After deduplication, we group the remaining generations by how they satisfied the positive constraints to encourage greater diversity in our knowledge base. 
Specifically, we group generations by the generated auxiliary verb, adverb of frequency, and comparative adjective and select only the generation with the best \score\footnotemark[5].
This further reduces the total number of generations to approximately 9\% of the overgenerated comparatives.

\subsubsection{Filtration by Contradiction}
\label{sec:filter-contradictions}

The tendency of language models to hallucinate information \cite{ji2022hallucination} sometimes results in unreliable generations which contradict each other. 
Using a RoBERTa contradiction classifier \cite{Liu2019RoBERTaAR,wang2022selfconsistency},
we discard comparatives that contradict others more often than not.
To increase the precision of the pre-trained classifier, we set a high threshold probability for classifying contradiction and entailment (0.99 and 0.85, resp.). 
Approximately 5\% of the overgenerated comparatives remain after this stage of filtering, from which we select only the $k=5$ best-scoring generations by their \score\footnotemark[5] for each entity pair.

\subsubsection{Discriminative Filtering}
\label{sec:knowledge-discriminator}

To further increase the quality of our retained knowledge, we build a final discriminator, adjustable to the desired balance between knowledge quantity and quality. 
We train a knowledge discriminator using crowdsourced annotations of valid and invalid knowledge generations, following prior work \cite{west2021symbolic}.
For a random 10K sample of our generated comparatives for unique entity pairs, we crowdsource the validity of each comparison (using the same setup described subsequently in \S\ref{sec:crowdsourced-evaluation}). 
We train a classifier to discriminate between aggregated ``Accept'' and ``Reject'' crowdsourced labels (see Appendix~\ref{app-discriminator} for additional details) and vary the threshold for the ``Accept'' class to filter at different levels of knowledge precision. 

\section{\neurocomps}
\label{sec:neurocomparatives}

Our large-scale generation effort produced 8.8m comparatives before discriminative filtering (details in App.~\ref{app:generation-details}), which we refer to as \neurocompsxl.
Specifically, \neurocompsxl includes 8.7M comparatives generated from GPT-2 XL, 78K from Llama-2-7b, 16.3K from InstructGPT, 10.7K from ChatGPT, and 6.6K from GPT-4.
We further apply our knowledge discriminator model on \neurocompsxl to create subsets containing only the top-50\% and top-20\% of comparatives, which we refer to as \neurocompsl and \ncsmall, respectively. 
While \neurocompsxl contains the greatest breadth of comparative knowledge, the more stringent filtering applied to produce \neurocompsl and \ncsmall results in the highest knowledge quality, without hurting diversity. 


The largest existing resource of comparative commonsense knowledge is WebChild \cite{tandon2017webchild}, collected via information extraction methods.
While WebChild contains over 18M general assertions covering 2M concepts and activities, we focus on its comparative knowledge, which spans 813K assertions over 335K entity pairs. 
Compared to WebChild, our \neurocompsxl corpus is 10x larger. 
Table~\ref{tab:qualitative} provides examples of \neurocompsxl in contrast to WebChild assertions across four pairs of entities. 
For ease of comparison, we convert the WebChild assertions from OpenIE triplets to a natural language format, similar to ours. 

The first set of examples for the entity pair (helicopters, planes) illustrates the more detailed, domain-specific properties, such as ``operating costs'', ``more cargo'', and ``stable in flight'' in \neurocompsxl. 
In contrast, WebChild assertions are more generic (e.g., ``cooler'', ``better'') and not specific to the domain of flight. 
This example also highlights how \neurocompsxl assertions are more informative and interesting to humans, as evidenced by their lower rate of rejection shown in Table~\ref{tab:qualitative}. 



We also compare our generated knowledge to  ATOMIC \cite{sap2019atomic} and ConceptNet \cite{speer2017conceptnet}. 
Although neither explicitly contains comparative knowledge, they do contain relations from which comparatives can be inferred. 
We use the \texttt{AtLocation} and \texttt{MadeUpOf} relations in ATOMIC, as well as the \texttt{AtLocation}, \texttt{PartOf}, and \texttt{MadeOf} relations in ConceptNet, to infer size comparisons over entities. 
We convert these entries to our \neurocompsshort  format for evaluation; for e.g., the ATOMIC triple (\texttt{human body}, \texttt{MadeUpOf}, \texttt{brain}) results in the comparative: ``Compared to brains, human bodies are larger.''.
We use human (\S\ref{sec:crowdsourced-evaluation}) and automatic (\S\ref{sec:diversity}) evaluation to compare the quality and diversity of \neurocomps with existing comparative knowledge resources.

\subsection{Human Evaluation of Validity} 
\label{sec:crowdsourced-evaluation}

We task 3 workers from Amazon Mechanical Turk with classifying each comparative into one of six categories: `True', `False', `Too subjective to judge', `Too vague to judge', `Too unfamiliar to judge' and `Invalid'.\footnote{This is an absolute evaluation scheme; relative comparisons of pairs of comparatives are somewhat unfair since the comparisons might be along different dimensions.}
We discard examples where there was no majority consensus among the 3 workers, and those marked as `Too unfamiliar to judge' by a majority vote.
Examples marked as `True' are considered valid, and all others, invalid. 
Appendix~\ref{app-crowdsourced-evaluation} details our annotation process (Fig.~\ref{fig:mechanical-turk-interface}). 
\begin{table}[ht!]
\centering
\resizebox{\columnwidth}{!}{%
\begin{tabular}{p{0.75cm} | l r r r r}
\toprule
& \textbf{Source} & \textbf{Size} & \textbf{Accept.} $\uparrow$ & \textbf{SB2} $\downarrow$ & \textbf{SB3} $\downarrow$ \\
\midrule
\multirow{3}{*}{\rotatebox[origin=c]{90}{\parbox{1.45cm}{\centering Existing KBs}}} & \HandPencilLeft \ ConceptNet & 34,355 & 91.8\% & 1.00 & 1.00\\
& \HandPencilLeft \ ATOMIC & 23,566 & 89.6\% & 1.00 & 1.00\\
& WebChild & 812,862 & 58.1\% & 0.77 & 0.71\\
\midrule[0.03em]
\multirow{3}{*}{\rotatebox[origin=c]{90}{\parbox{1.45cm}{\centering Few-shot LLMs}}} & InstructGPT & - & 72.7\% & 0.91 & 0.89\\
& ChatGPT & - & 86.2\% & 0.90 & 0.88\\
& GPT-4 & - & 89.4\% & 0.87 & 0.84\\
\midrule[0.03em]
\multirow{3}{*}{\rotatebox[origin=c]{90}{\parbox{1.45cm}{\centering \neurocompsshort (GPT-2)}}} & \neurocompsxl & 8,709,810 & 76.9\% & 0.64 & 0.58\\
& \neurocompsl & 4,354,905 & 84.4\% & 0.64 & 0.58\\
& \ncsmall & 1,741,962 & 90.1\% & 0.65 & 0.59\\
\midrule
\multirow{4}{*}{\rotatebox[origin=c]{90}{\parbox{1.9cm}{\centering \neurocompsshort (Other LLMs)}}}& \neurocompsshort(LLaMA-7b) & 77,798 & 90.0\% & 0.64 & 0.59\\
& \neurocompsshort(InstructGPT) & 16,300 & 83.8\% & 0.60 & 0.56 \\
& \neurocompsshort(ChatGPT) & 10,756 & 88.6\% & 0.48 & 0.44 \\
& \neurocompsshort(GPT-4) & 6,630 & 89.0\% & 0.52 & 0.48 \\
\bottomrule
\end{tabular}
}
\caption{
Size, human acceptance rates (on a 0.5k subset) and diversity measures, Self-BLEU-2 (SB2) and Self-BLEU-3 (SB3) for different comparatives. 
\HandPencilLeft \ indicates human-authored comparatives.
\neurocompsshort generated from GPT-2 achieve a better trade-off between acceptance and diversity than generations from few-shot LLMs (no filtering); our filtered \neurocompsshort from LLMs fare even better.
Discriminative filtering of \neurocomps improves acceptance without hurting diversity.
}
\label{tab:human-eval-results}
\end{table}

We evaluate 500 randomly sampled comparatives from \neurocompsxl, WebChild, ConceptNet, and ATOMIC. 
For evaluation of comparative knowledge extracted from GPT-3, ChatGPT, and GPT-4 without our framework (i.e., without any filtering), we obtain a sample of 500 completions to the same prompts used to generate the sampled \neurocomps (see Appendix~\ref{app-gpt3} for additional discussion on the impact of filtering on ChatGPT and GPT-4).
Human acceptance results are shown in Table~\ref{tab:human-eval-results} along with the size (total num. of comparatives) of different sources of comparative knowledge. 
While human-authored comparatives in ConceptNet and ATOMIC have the highest acceptance, these sources are the smallest in size, involved expensive human efforts and cannot be arbitrarily scaled.
Among generated comparatives, \neurocompsxl achieves nearly a $20\%$ absolute improvement in human acceptance relative to WebChild, 
while containing over 10x more comparative knowledge. 
On average, few-shot prompting without filtering achieves lower acceptance than our \neurocompsshort with the same LLMs, highlighting the benefits of our approach.
\neurocompsshort generated by LLaMA-7b achieved a similar human acceptance rate as that generated by GPT-2, suggesting that high-quality knowledge may be acquired from smaller-scale LMs. 
\paragraph{Filtering Contradictions improves \neurocomps}
\label{sec:filtering-impact}
We conduct an ablation to study the impact of filtering contradictions on generating \neurocomps (\S\ref{sec:filter-contradictions}).
We obtain a sample of comparatives from GPT-2-XL without the contradiction filter. 
The overall acceptance rate of these comparatives is 69.1\%, which is an absolute decrease of 7.8\% compared to \neurocompsxl, confirming the importance of contradiction filtration. 

\begin{figure}[t!]
    \centering
    \includegraphics[width=\columnwidth]{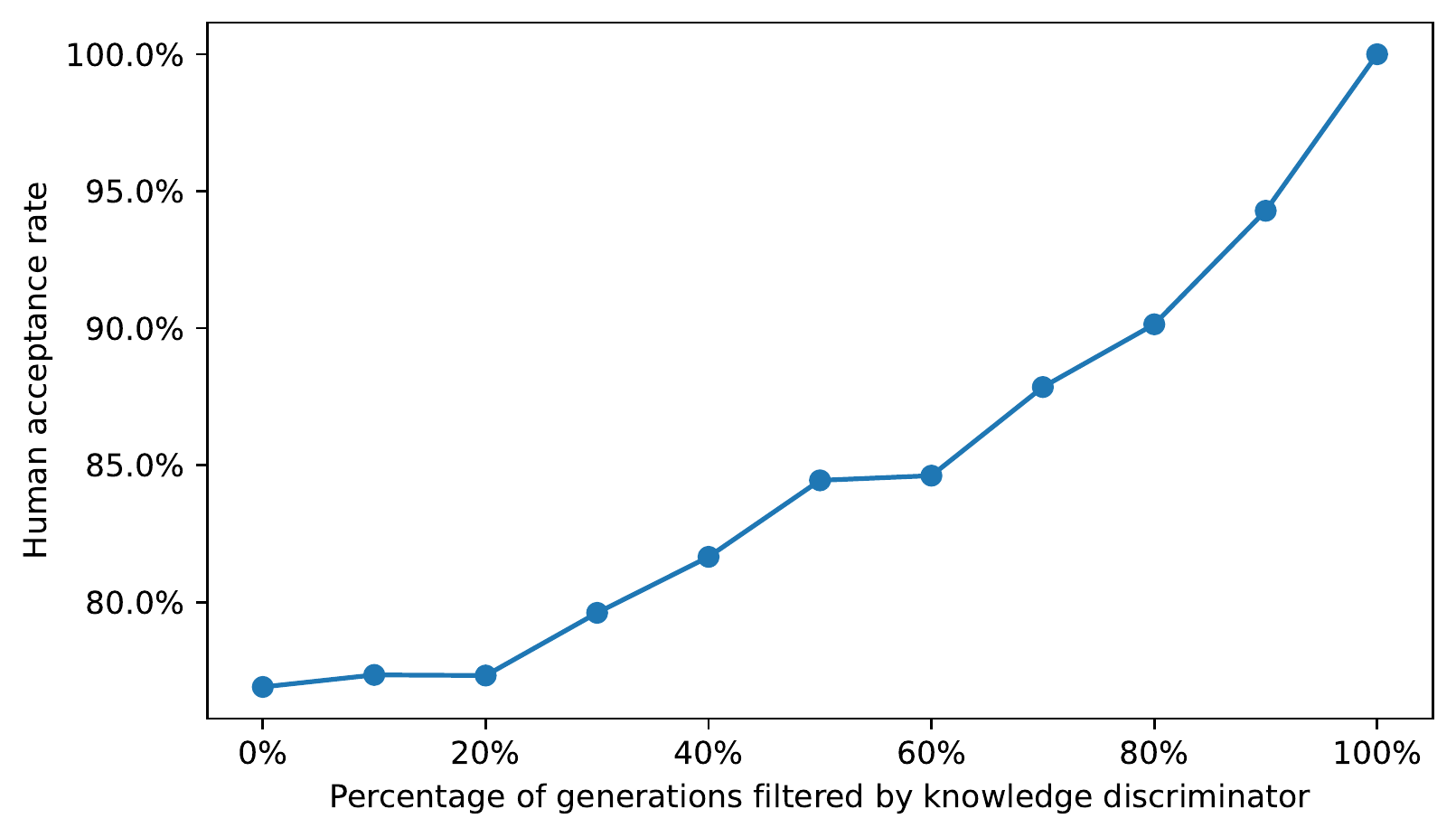}
    \caption{As our knowledge discriminator 
    gets stricter, human acceptance of \neurocomps increases. 
    }
    \label{fig:acceptance-by-percent-filtered}
\end{figure}

\paragraph{Impact of Discriminative Filtering}
Figure~\ref{fig:acceptance-by-percent-filtered} and Table~\ref{tab:human-eval-results} show that \neurocompsl, which includes only the top-50\% of our generated knowledge, achieves an acceptance rate of 84.4\%, a 7.5\% absolute increase relative to \neurocompsxl. 
The top-20\% filtering used to produce \ncsmall improves the acceptance rate even further to 90.1\%, which exceeds the acceptance of few-shot prompting with both ChatGPT and GPT-4 (without our filtering) and is on par with human-authored sources. 
At this level of filtering, \ncsmall is still over 2x larger than WebChild while having an approximately 32\% absolute gain in human acceptance rate. 



\subsection{\neurocomps' Diversity}
\label{sec:diversity}

We report Self-BLEU \cite{zhu2018texygen}, using both bigrams (SB2) and trigrams (SB3), to compare the diversity of comparatives from different sources in \autoref{tab:human-eval-results}. 
We randomly sample 500 entity pairs from WebChild containing at least 5 comparisons. 
For few-shot prompting with GPT-3, ChatGPT, and GPT-4, we use the same prompts (see \S\ref{sec:crowdsourced-evaluation}) to obtain the top-5 generations for each entity pair.
For each source, we compute Self-BLEU among the 5 candidates.
Since the comparatives from ConceptNet and ATOMIC are limited to a single relation (size), measure of diversity doesn't apply.
\neurocompsxl and \neurocompsl exhibit the greatest diversity, with a 31\% reduction in Self-BLEU-3 relative to few-shot prompting with GPT-4. 
Crucially, knowledge discrimination does not impact the diversity of \ncsmall.


While \neurocompsxl by design contain 5 comparisons for each pair of entities, the amount of comparative knowledge per entity pair in WebChild is heavily skewed: approximately 80\% of the entity pairs have only 1 comparison and over 10k assertions comparing the entities ``man'' and ``woman.''
WebChild is also heavily skewed towards a small number of frequently-occurring relations (e.g., ``better''). 
We illustrate the skew in \autoref{fig:adj-frequency} in Appendix~\ref{app-diversity}.

To further quantify the diversity of comparatives in \neurocompsxl and WebChild, we consider identical relations from each source by considering unique comparative adjective phrases.
We build a frequency-based probability distribution for relations in each source, and use it to compute their respective entropies--- higher entropy indicates greater diversity of comparative relations. 
Our results show that the entropy of \neurocompsxl is 7.9, which is 30\% higher (indicating greater diversity) than the WebChild entropy of 6.1. 




\section{Downstream Task Performance}
\label{sec-downstream}

While \neurocomps are demonstratably diverse (\S\ref{sec:diversity}), the impact on downstream task can be measured via their coverage on existing benchmarks. 
Here, we demonstrate \neurocompsshort' wide knowledge coverage without compromising knowledge quality on three benchmarks.

To this end, we use two existing benchmarks: Elephant \cite{elazar2019large} and Verb Physics Reannotated (\citealp[VPR]{elazar2019large}).
Elephant contains 486 comparisons of sizes of various transportation vehicles and animals.
VPR reannotates VerbPhysics  \citet{Forbes2017VerbPR} by filtering out the examples where the objects were not comparable, or the annotators disagreed among themselves.  
This re-annotation results in 2964 examples of object relations (1927 unique pairs of objects) along the dimensions of speed, length and mass. 

The unstructured knowledge in \neurocomps facilitates easy integration with LLMs through natural language, therefore enabling the use of comparative reasoning abilities of LLMs. 
As there are no benchmarks for systematically evaluating the comparative reasoning capabilities, we constructed a new comparative reasoning QA benchmark, \textbf{ComparativeQA} to investigate this question. 
Specifically, we identified all comparatives that were unanimously assigned a label of `True' by annotators in our human evaluation of validity experiments (\S\ref{sec:crowdsourced-evaluation}). 
We then rephrased these comparatives as questions in order to construct a QA benchmark (see App.~\ref{app-comp-reasoning-qa}). 
This produced a total of 656 comparative reasoning questions.

Finally, the COPEN dataset \cite{peng2022copen} is a benchmark for probing conceptual knowledge in pre-trained language models. Two tasks from COPEN (Conceptual Similarity Judgement and Conceptualization in Contexts) are related to our task of acquiring comparative knowledge. Therefore, we evaluate the benefit of our framework on these two additional downstream tasks.



\subsection{Coverage on Elephant}
\label{sec:downstream}

We evaluated five different versions of \neurocomps that differed by model scale: \ncsmall generated from GPT-2 XL, \neurocompsshort from LLaMA-2-7b, from GPT-3, from ChatGPT, and from GPT-4; all variants went through discriminative filtering.
Because \neurocomps are generated without restrictions on the attribute used for comparison, only a subset of these comparisons are along the size dimension corresponding to the 486 comparisons in Elephant. 
We identify this subset via a simple string matching approach and filter out the remaining generations for this evaluation. 

As shown in Table \ref{tab:elephant-vpr-eval-results}, GPT-4 \neurocompsshort achieved nearly perfect exact match accuracy---$98.7\%$ express the correct size relationship according to the Elephant annotations, and higher overlap with the Elephant dataset. 
We observe a very clear trend that more capable models generate \neurocompsshort with higher quality. 


\subsection{Coverage on VPR}
\begin{table}
\centering
\resizebox{0.95\columnwidth}{!}{%
\begin{tabular}{l  r r  r r}
\toprule
& \multicolumn{2}{c}{\textbf{Elephant}}  & \textbf{VPR} & \\
\cmidrule(lr){2-3}\cmidrule(lr){4-5}
\textbf{\neurocompsshort Source} & \textbf{\#Overlap} & \textbf{Acc. } $\uparrow$ & \textbf{\#Overlap} & \textbf{Acc. } $\uparrow$ \\
\midrule
\ncsmall (GPT-2 XL) & 205 & 66.7\% & 869 & 66.7\% \\
\neurocompsshort (LLaMA-2-7b) & 132 & 90.2\% & 760 & 71.2\% \\
\neurocompsshort (InstructGPT) & 250 & 76.8\% & 1106 & 76.2\% \\
\neurocompsshort (ChatGPT) & 189 & 97.9\% & 1090 & 88.2\% \\
\neurocompsshort (GPT-4) & 226 & \textbf{98.7}\% & 1351 & \textbf{93.3\%} \\
\bottomrule
\end{tabular}
}
\caption{Coverage (overlap) and accuracy of \neurocompsshort on 486 examples in the Elephant dataset as well as 2964 examples in the VPR dataset.
}
\label{tab:elephant-vpr-eval-results}
\end{table}
We used our framework to generate \neurocompsshort with the unique object pairs in VPR and additionally filtered them to contain comparisons only along speed, length, and mass using few-shot classification with Falcon-40b-instruct. 
We then used a BART-large model trained on MNLI to determine if a pair of \neurocompsshort and VPR entail each other. 
Entailment means that our generated knowledge agrees with VPR used as ground truth (see Appendix \ref{app-verb-physics-reannotated} for details). 

In Table \ref{tab:elephant-vpr-eval-results}, we can see that GPT-4 \neurocompsshort again achieved the best overall accuracy of $93.3\%$ as well as the highest coverage, highlighting the value of large-scale models. 
On top of being accurate, our \neurocompsshort generates much more diverse knowledge beyond these three evaluated attributes, not directly reflected in the results. 

\subsection{Results: ComparativeQA}
\label{sec:qa}
\begin{table}
\centering
\begin{tabular}{l r}
\toprule
\textbf{LLaMA-2-7b Finetuning Source} & \textbf{Acc.} $\uparrow$ \\
\midrule
None (5-shot) & 60.7\% \\
\midrule[0.03em]
\ncsmall (GPT-2) & \textbf{91.0\%} \\
\ncsmall (GPT-2 + ChatGPT + GPT-4) & \textbf{92.7\%} \\
\midrule[0.03em]
Reversed-\ncsmall & 14.6\% \\
WebChild  & 79.0\% \\
\bottomrule
\end{tabular}
\caption{LLaMA-2-7b performance on comparative reasoning QA. fine-tuned on 50k sampled comparatives from various sources. All results are 5-shot. 
}
\label{tab:qa-eval-results}
\end{table}

We used a LLaMA-2-7b model for QA on our constructed ComparativeQA benchmark.
As shown in Table \ref{tab:qa-eval-results}, vanilla LLaMA-2-7b achieves only 60.7\% accuracy on this dataset, using a prompt with five in-context examples. 
After finetuning LLaMA-2-7b for 3 epochs on a separate dataset of 50k QA comparatives sampled from \ncsmall from GPT-2, its accuracy on the test dataset increases to 91\%. 
This is despite the lack of overlap between the entity pairs in the 50k \neurocompsshort sampled for training and the test set. 
On additionally including 17k \ncsmall from ChatGPT and GPT-4 for finetuning, we see a further increase in validation accuracy to $92.7\%$.

As a control experiment, we finetuned on directionally reversed comparatives on the sampled \ncsmall.
Finetuning on this set of incorrect comparatives degrades the accuracy of LLaMA-2-7b on the test set to 14.6\%, highlighting the quality of our \neurocompsshort.
In addition, we also finetuned LLaMA-2-7b using a random sample of 50k comparatives from WebChild, and this results in an accuracy of 79\%.

\subsection{Results: COPEN Dataset}

The Concept Similarity Judgment (CSJ) task from the COPEN dataset is related to a component of our framework which aims to identify entity pairs sharing common attributes. Specifically, our comparative prompt construction and filtering method (Section~\ref{sec:prompt-and-constraints}) implicitly performs a concept similarity judgment between pairs of entities. We therefore applied this approach to query and candidate entities in the CSJ test set, using our GPT2-XL length-penalized perplexity score of comparative prompts to identify the best candidate which matches the concept corresponding to each query.

In the zero-shot setting, we found that our approach produces a 19\% relative improvement in the performance of GPT-2-XL on the development set (increasing from 11.3 to 13.44) and a 14\% relative improvement on the test set (increasing from 11.7 to 13.3). Note that these results were obtained without any additional training of the model.

We also found that the Conceptualization in Contexts (CiC) task from the COPEN dataset can benefit from our approach. This task requires the model to identify the best concept for an entity in a context. The baseline performance of GPT-3.5-turbo-1106 in the zero-shot setting for this task is 40.9\%. We used GPT-2-XL to generate \neurocomps among candidate concepts and used them as additional context in the prompt, which improved the accuracy of GPT-3.5-turbo-1106 to 42.8\%. These results demonstrate the utility of our \neurocomps KB and knowledge acquisition framework for improving performance on downstream tasks without the need for additional model training.

\section{Related Work}
\label{sec:related}

\textbf{Comparative Knowledge:} Despite the significance of comparative knowledge \cite{hofstader2013surfaces}, resources for the same are few and far between.
Those that do exist have been collected almost exclusively via IE and data mining methods \cite[i.a.]{Jindal2006MiningCS, Cao2010ExtractingCC,Jain2011HowDT,Jang2012PredictiveMO,Tandon2014AcquiringCC,tandon2017webchild,elazar2019large}, raising questions about coverage and diversity.
Our approach is designed to address such issues via distilling knowledge from models at different scales.
Most modern knowledge resources such as ATOMIC \cite{sap2019atomic}, VerbPhysics \cite{Forbes2017VerbPR} and DoQ \cite{elazar2019large} may contain implicit comparisons via relationships between and properties of entities.
A few knowledge resources involve explicit comparisons, but only along a few specific dimensions like physical properties (e.g. size \cite{Bagherinezhad2016Elephants} or number \cite{narisawa-etal-2013-204}).
Recently, \citet{yu2023pretraining} collect data from structured and unstructured sources on specific real-world entities.
\citet{shivade-etal-2015-corpus} compare gradable lexical items, primarily adjectives and adverbs.
In contrast, our \neurocompsshort involve \textit{explicit} comparisons between \textit{nominal everyday concepts}, \textit{without restrictions on the comparison dimensions}.

\noindent\textbf{LM Knowledge Distillation:} The ascendance of LLMs has begun to replace expensive, manually constructed knowledge bases due to their coverage benefits \cite{alkhamissi2022review}. 
LLMs have been used to create resources of factual knowledge \cite{petroni2019language}, structured knowledge graphs \cite{hao2022bertnet}, instructions for further training \cite{wang2023selfinstruct} and training data for different tasks \cite[\textit{i.a.}]{liu-etal-2022-wanli,chakrabarty-etal-2022-flute}.
Our overgenerate and filter framework is inspired by \citet{west2021symbolic} who distill GPT-3  into a commonsense KG with a supervised critic.
Our \neurocompsshort focuses on \textit{comparative knowledge} distilled from (among others) GPT-2 and LLaMA-2 with neuro-symbolic constrained decoding; \citet{allaway2022penguins} and \citet{Bhagavatula2022I2D2} use a related method to distill generics knowledge \cite{hampton2012generics} from GPT-2. Beyond knowledge distillation, neuro-symbolic constrained deocding with \neurologic has been applied to tasks such as counterfactual generation \cite{howard2022neurocounterfactuals} and prompt engineering \cite{rosenman2023neuroprompts}.

\section{Conclusion}
\label{sec:discussion}

We demonstrate distillation of high-quality comparative knowledge from LMs at different scales and produce \neurocomps: the largest comparative knowledge corpus to date. 
\neurocompsxl is 10x larger, 30\% more diverse, and 19\% more human acceptable than existing sources; with  knowledge discrimination, we additionally achieve over 90\% human acceptance. 
Our work highlights the value of comparative knowledge and exploits both neuro-symbolic manipulation of small-scale models and extreme-scale models for knowledge distillation.
In our evaluations which demonstrated the utility of \neurocomps for downstream tasks, we primarily focused on comparative reasoning since it is most directly related to the knowledge we acquired. However, a promising direction for future research would be investigating other downstream tasks which can benefit from training on \neurocomps.

\section*{Limitations}

\neurocomps is a collection of fully generated data with limited manual verification. 
Caution must be exercised around training and deploying models on such data, due to reasons outlined below.

While our work centers around distilling knowledge from language models, it is well known that language models generate misinformation as well as toxic content.
The scale of generations in our paper makes it challenging to manually analyze each generation.
We expect that our filtering stage (\S\ref{sec:filtering}) and knowledge discriminator (\S\ref{sec:knowledge-discriminator}) are able to filter out many contradictory statements, but the veracity of the filtered remainder is challenging to determine.
It is conceivable that there remain some fallacies in the data.
As our comparisons are designed to be restricted to be between physical objects (as our root seed entities), we avoid comparisons between animate entities and any toxic content that might be associated with such comparisons.

\neurocomps is a resource in English only.
Further, we restricted our entities to be objects in the real world which are nouns. 
However, there could be many potentially useful comparisons among verbs and adjectives. 
Due to limited resources, we leave the investigation of those to future work.

\bibliography{custom}

\appendix
\section{Additional Details on the Generation of \neurocomps}
\label{app:generation-details}

\subsection{\neurologic Background}
\label{app:neurologic}

\neurologic accepts a series of constraints $D(\mathbf{a}, \mathbf{y})$ which are true iff `$\mathbf{a}$' appears in the generated sequence `$\mathbf{y}$'. where each constraint is a set of \textit{clauses} $\{C_{i} \mid i \in 1, \cdots m\}$ consisting of one or more predicates in Conjunctive Normal Form (CNF):
\vspace*{-2mm}
\begin{equation}
\label{eq:cnf}
    \underbrace{(D_{1} \lor D_{2} \cdots \lor D_{i})}_\textrm{$C_1$} \land \cdots \land \underbrace{(D_{k} \lor D_{k+1} \cdots \lor D_{n})}_\textrm{$C_m$}.
\end{equation}
Each constraint $D$ might be positive or negative; $D(\mathbf{a}_i, \mathbf{y})$ is satisfied (i.e., evaluates as true) if $\mathbf{a}_i$ is present or absent, respectively, in $\mathbf{y}$

\neurologic employs a beam search approximation of an objective function which maximizes the probability of the generated sequence while penalizing deviations from $m$ clauses:
\vspace*{-2mm}
\begin{equation}
    \hat{\mathbf{y}_t} = \text{arg} \max_{\mathbf{y} \in \mathcal{Y}} p_{\theta}(\mathbf{y}_t | \mathbf{y}_{< t}) - \lambda \sum_{j=1}^{m} (1 - C_{j})
    \label{eq:objective}
\end{equation}
where $\lambda \gg 0$ penalizes deviations from the constraints. 

Candidates are scored at each $t$ per their (partial) satisfaction of the constraints:
\vspace*{-2mm}
\begin{equation}
    f(\mathbf{y}_{\le t}) = \log p_{\theta} (\mathbf{y}_{\le t} | \mathbf{x}) + \lambda \max_{D(\mathbf{a}, \mathbf{y}_{\le t})} \frac{|\hat{\mathbf{a}}|}{|\mathbf{a}|}
    \label{eq:scoring}
\end{equation}
where $\hat{\mathbf{a}}$ represents a subsequence of $\mathbf{a}$ in the current generation. 
This has the effect of preferring candidates which at least partially satisfy multi-token constraints; for example, a generated sequence $\mathbf{y}_{\le t} =$ {``Compared to train tickets, airline tickets are generally more''} would be rewarded for partially satisfying the constraint $\mathbf{a} =$ {``more expensive''} via its subsequence $\hat{\mathbf{a}} =$ {``more''}. 

Unlike the top-$k$ selection strategy used in traditional beam search, \neurologic performs pruning, grouping, and selection steps to identify the best candidates which satisfy the given constraints. Specifically, candidates which irreversibly violate one or more constraints are pruned, and the remaining candidates are grouped according to their number of satisfied clauses in order to encourage diversity. The best candidate within each group is then selected according to the scoring function in Equation~\ref{eq:scoring}.

Each pass of \neurologic returns multiple generations, which are scored according to the sum of their length-penalized log probabilities:
\vspace*{-2mm}
\begin{equation*}
    \frac{1}{N^{\alpha}} \sum_{t=1}^{N} \log p_{\theta} (\mathbf{y}_{t} | \mathbf{y}_{< t}) 
    \label{eq:neurologic_score}
\end{equation*}
\noindent
where $N$ denotes the length of the generated sequence $\mathbf{y}$ and $\alpha$ is a length penalty to encourage shorter generations (we use $\alpha = 0.1$). We refer to this score as as the \score.

\subsection{Constraint Sets for \neurologic}
\label{app:constraints}

Table~\ref{tab:positive-constraints} provides the positive constraints used in \neurologic decoding.
The table lists tokens used for two different positive constraint sets. 
For each of the 30 pairwise combinations of these auxiliary verbs and adverbs, we generate a completion of the prompt where the corresponding auxiliary verb and adverb is required to be present in the generation. 
\begin{table}
\centering
\small
\begin{tabular}{l l}
\toprule
\textbf{Auxiliary verbs}&  \textbf{Adverbs of frequency} \\
\midrule
have & typically \\
need & often \\
may & always \\
are & generally \\
would & normally \\
\bottomrule
\end{tabular}
\caption{Positive constraint sets. \swabha{I think may is a modal, also check if "need" counts as an auxiliary verb...}}
\label{tab:positive-constraints}
\end{table}

\begin{table}[ht]
\centering
\small
\resizebox{0.5\textwidth}{!}{%
\begin{tabular}{l l l }
\toprule
\textbf{Prompt} & \textbf{Aux. Verb}&  \textbf{Adverb} \\
\midrule
Compared to cherries, peaches \ldots & have & typically \\
Compared to cherries, peaches \ldots & have & often \\
Compared to cherries, peaches \ldots & have & always \\
\vdots & \vdots & \vdots \\
Compared to cherries, peaches \ldots & would & normally \\
\bottomrule
\end{tabular}
}
\caption{Example of the prompt and 30 combinations of positive constraints for the entity pair $(\texttt{cherries}, \texttt{peaches})$.}
\label{tab:prompt-constraints-example}
\end{table}
An illustration of the prompt and the positive constraint combinations used to generate comparisons for an entity pair is provided in Table~\ref{tab:prompt-constraints-example}.

Table~\ref{tab:positive-constraints} provides the negative constraints used in \neurologic decoding.

We use GPT-2 XL as our language model, which has 1,542M parameters. For decoding with \neurologic, we use a beam size of 15, length penalty of 0.1, and an $n$-gram size of 3 for preventing repetitions. We use $\beta = 1.25$ as the reward factor for in-progress constraint satisfaction and set the constraint satisfaction tolerance to 3, which means that only candidates which have a number of satisfied constraints within 3 of the maximum are kept at each step. The hyperparameters are manually curated. Please refer to \citet{lu2021neurologic} for details on these hyperparameters.

Our experiments were conducted on a cluster with Nvidia RTX A6000 GPUs. We distributed the generation across 64 GPUs, with each GPU running 4 decoding iterations in parallel. The total compute time to generate our knowledge base in this environment was approximately 5 weeks.  

\subsection{Ordered Generation}
\begin{figure}[t]
    \centering
    \includegraphics[width=0.95\columnwidth]{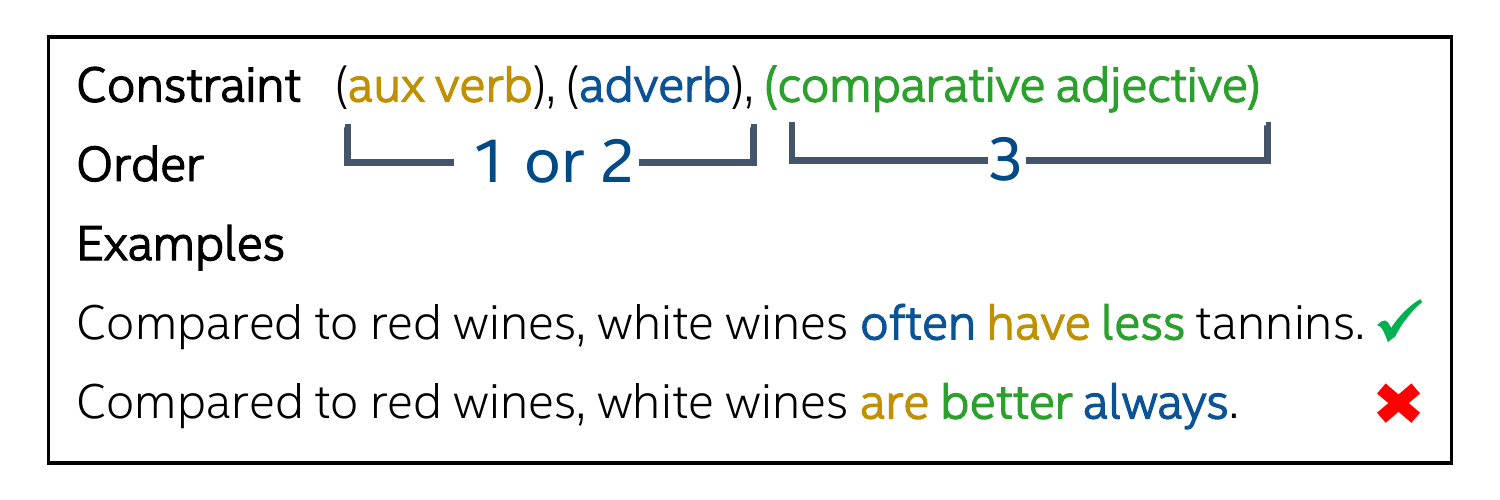}
    \vspace*{-2mm}
    \caption{Examples of generated comparatives which satisfy and violate our constraint ordering.
    }
    \label{fig:ordered-constraint-satisfaction}
\end{figure}

We additionally modify \neurologic to handle ordered constraint satisfaction, for fine-grained control.
For each clause $C_{i}$, we assign one or more order indices $m_{i} \in \{1, ..., m\}$ which correspond to the positional order in which clause $C_{i}$ can appear in the generation. Specifying more than one order index allows a clause to appear in multiple different positions. 
Ordered constraint satisfaction provides more fine-grained control for generating valid comparatives, as illustrated in Figure~\ref{fig:ordered-constraint-satisfaction}.

\section{Details of knowledge discriminator model}
\label{app-discriminator}

We use 80\% of the labeled data for training the knowledge discriminator and 20\% for validation.
We trained the knowledge discriminator on a Ubuntu 18.04 system with a single Nvidia RTX 3090 GPU. Specifically, we finetune RoBERTa-large  previously trained  on MNLI\footnote{\url{https://huggingface.co/roberta-large-mnli}} using a learning rate of 5e-6, a batch size of 32, and a dropout probability of 0.1. Hyperparameters are manually curated.
We train the model for a maximum of 50 epochs and monitor precision at $\textrm{recall} = 80\%$ on the validation set, terminating training if this metric fails to improve for 5 consecutive epochs. 
The total training time of the model was 13 minutes.
Precision and recall on the validation set were 0.589 and 0.642, respectively.

\section{Details of experiments with InstructGPT, ChatGPT, and GPT-4}
\label{app-gpt3}

To compare our knowledge generations to InstructGPT, ChatGPT, and GPT-4, we use a prompt which instructs each model to complete a statement comparing two entities. The instruction is followed by five hand-crafted examples and the prefix that we want the model to complete in order to form a comparative knowledge statement. An example of the full prompt used to generate a comparative knowledge statement for the entity pair (computer keyboards, game controllers) is provided below.
\singlespacing
\begin{addmargin}[1em]{1em}
\small{
\textsl{Complete a statement which compares two entities.\newline
Compared to blueberries, pineapples are heavier.\newline
Compared to chairs, sofas are larger.\newline
Compared to salad, pizza is less healthy.\newline
Compared to a knife, a machete is more dangerious.\newline
Compared to a bicycle, a skateboard is slower.\newline
Compared to computer keyboards, game controllers
}}
\end{addmargin}
\singlespacing
We use OpenAI's API with the text-davinci-001 model for InstructGPT, gpt-3.5-turbo-0613 for ChatGPT, and gpt-4-0613 for GPT-4 results. We use the default parameter settings for each model and evaluate human acceptance using the first generation returned for each prompt. For diversity evaluations, we utilize the first 5 generations returned by the API for each prompt. 


\section{More on Evaluation}
\label{app-eval}

\subsection{More on diversity of \neurocomps}
\label{app-diversity}

\begin{figure*}
     \centering
     \begin{subfigure}[b]{0.49\textwidth}
         \includegraphics[height=\textwidth]{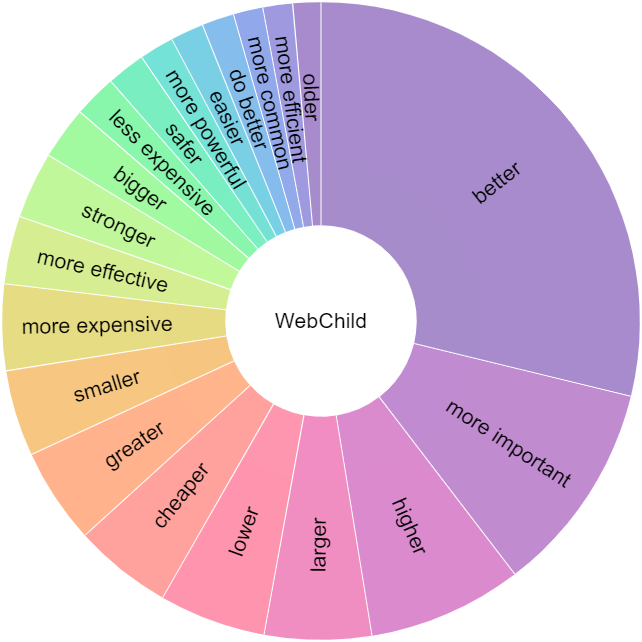}
     \end{subfigure}
     \begin{subfigure}[b]{0.49\textwidth}
         \includegraphics[height=\textwidth]{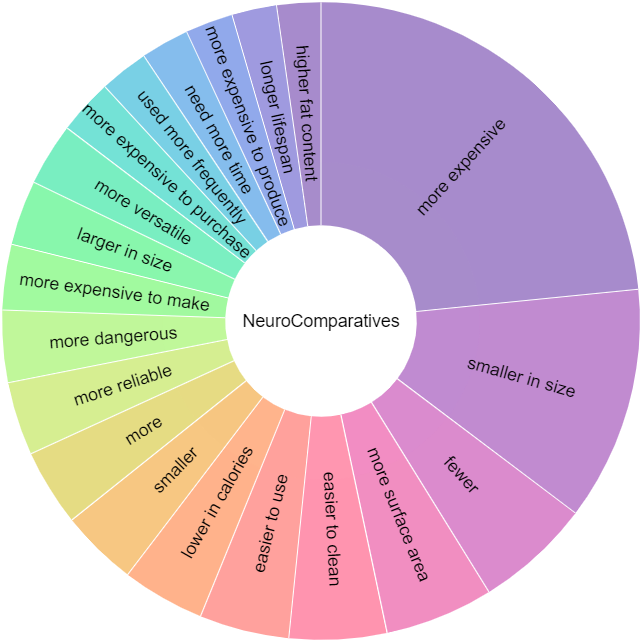}
     \end{subfigure}
     \hfill
     \caption{The distribution of the top-20 relations in WebChild is more skewed than \neurocompsxl. }
    \label{fig:adj-frequency}
\end{figure*}

Figure~\ref{fig:adj-frequency} depicts the top-20 most frequent relations in each source, showing that the WebChild relations are more skewed, with its most-frequent relation (``better'') representing over 12\% of all relations. 
In contrast, the most frequent relation in \neurocompsxl (``more expensive'') represents only 4\% of all relations. 
The most-frequent relations in \neurocomps are also generally more descriptive and less subjective than those in WebChild. 

\subsection{Crowdsourced evaluation details}
\label{app-crowdsourced-evaluation}

\begin{figure}[h]
    \centering
    \includegraphics[width=1\columnwidth]{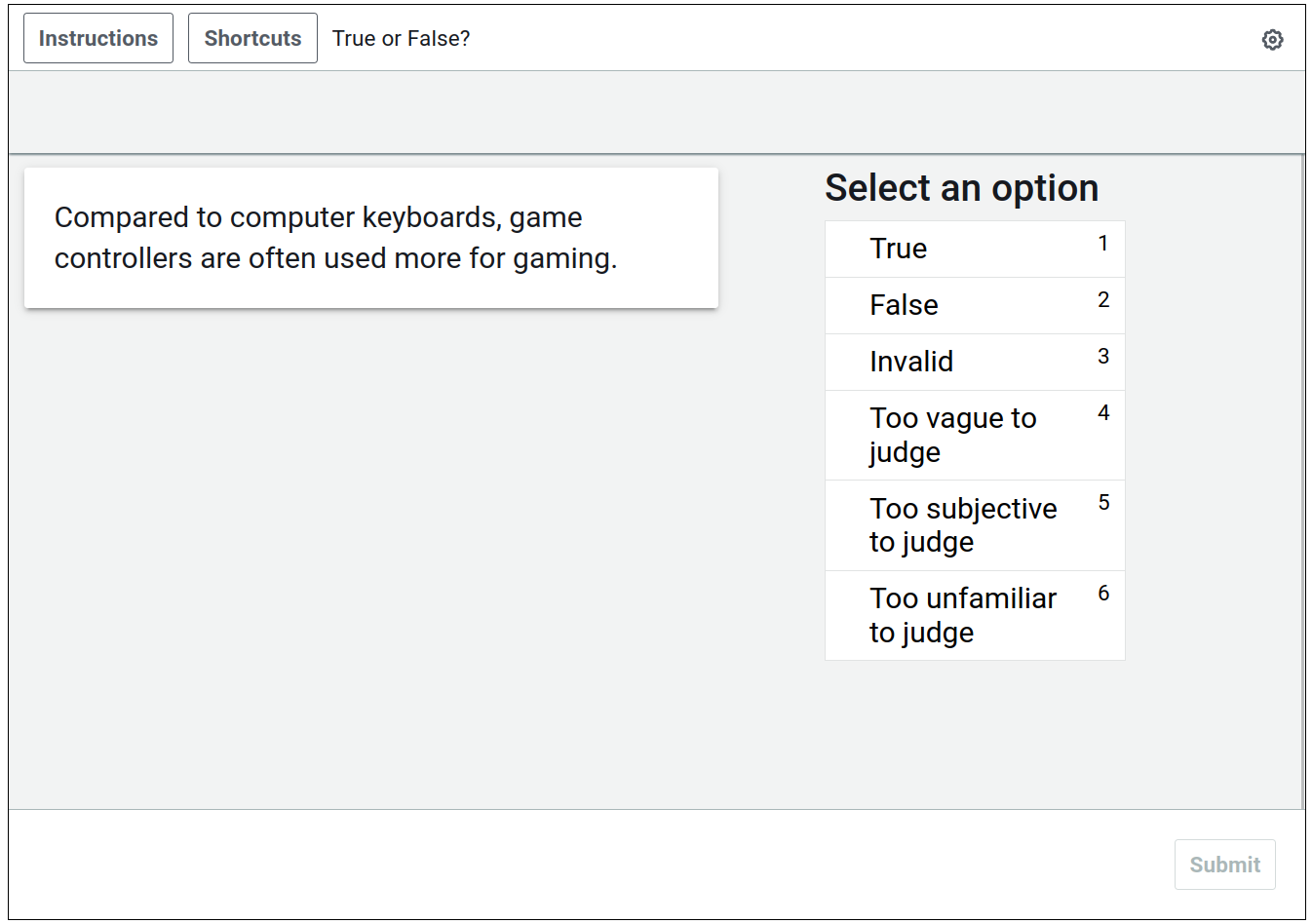}
    \caption{Validity labeling interface for crowdsourced workers}
    \label{fig:mechanical-turk-interface}
\end{figure}

Our crowdsourced evaluations utilized Amazon Mechanical Turk workers who were required to have completed at least 5,000 HITs, have a lifetime task acceptance rate $\ge 95\%$, and have achieved the \lq Masters' qualification. A reward of \$0.07 was paid to the workers for each submitted label. 

To ensure that all sources of knowledge were evaluated in the same form, we transformed triples in WebChild into a comparative knowledge statement format. Specifically, we pluralized the head and tail entities of each triple using the \texttt{inflect} Python package and then formed a comparative knowledge statement using the following template:  ``Compared to \{tail\}, \{head\} \{relation\}''.

We provided the following set of instructions and examples to the workers. 

\subsection{Instructions}

In this task, you will be given a sentence which compares two entities.

\begin{itemize}
    \item Determine whether the comparison is true or false (or indicate that you cannot determine its truthfulness) by selecting one of the 6 options.
    \item If the sentence is incoherent or not a valid comparison, select "Invalid". Please be forgiving of spelling or grammatical errors and avoid labeling it as invalid if the sentence only has minor grammatical mistakes.
    \item If the comparison is too vague or requires additional information to determine its truthfulness, select "Too vague to judge".
    \item If the comparison is overly subjective or expresses a personal opinion which is not commonly held by most people, select "Too subjective to judge".
    \item If the terms are too obscure or you do not know the truth of the comparison, it is okay to select "Too unfamiliar to judge". If you can answer (e.g., based on likelihood), please provide a response.
    \item If a comparison in unjudgeable due to more than one of the above reasons, select the option corresponding to the primary reason it cannot be judged.
\end{itemize}

\subsection{Examples}
\label{app:examples}

\textbf{True}: "Compared to homes, office buildings are more expensive to build."
\singlespacing \noindent 
\textbf{False}: "Compared to doctorates, master's degrees are more difficult to obtain."
\singlespacing \noindent 
\textbf{Invalid}: "Compared to toothbrushes, utility knives may be less efficient at cleaning always on."

\noindent Explanation: It is unclear what being "less efficient at cleaning always on" means.
\singlespacing \noindent 
\textbf{Too vague to judge}: "Compared to text messages, video chats generally have higher levels."

\noindent Explanation: Higher levels of what? The comparison lacks details needed to determine its truthfulness.
\singlespacing \noindent 
\textbf{Too subjective to judge}: "Compared to french toast, pancakes are better."

\noindent Explanation: Although this comparison may be true for many people, it is a subjective opinion which varies substantially from person-to-person.
\singlespacing \noindent 
\textbf{True}: "Compared to frozen foods, fresh foods are healthier."

\noindent Explanation: While this comparison could also be considered an opinion, it is one which is widely held by most people and therefore should be labeled as True.
\singlespacing \noindent 
\textbf{Too unfamiliar to judge}: "Compared to gyroscopes, microelectromechanical systems may often provide better performance."

\noindent Explanation: I am too unfamiliar with "gyroscopes" and "microelectromechanical systems" to judge this comparison.

\section{Case Study}
\subsection{Verb Physics}
\label{app-verb-physics-reannotated}

\paragraph{Filtering}
Here is the prompt used for filtering:
\singlespacing
\begin{addmargin}[1em]{1em}
\small{
\textsl{
Solve a textual classification task by having a Thought, then Finish with your answer. Thought can reason about the current situation. Finish[answer] returns the answer and finishes the task.
There are 4 classes you need to decide among speed, length, mass, and others. Don't answer with anything else. Here are some examples: \newline
\{FEW-SHOT EXAMPLES\} \newline (END OF EXAMPLES)\newline 
Sentence:
}}
\end{addmargin}
\singlespacing
Some examples are manually labeled and annotated and put in the place of \{FEW-SHOT EXAMPLES\} above(refer to Tab.~\ref{tab:fewshot-classification-example} for a complete list of few-shot examples). We make sure those examples are not in the final evaluation set. 

\paragraph{Classification}
We then used a BART-large model trained on MNLI\footnote{\url{https://huggingface.co/facebook/bart-large-mnli}} to determine if a pair of \neurocomps and VPR entail each other. 
We perform this entailment classification twice---first using \neurocomps as the premise and VPR examples as hypothesis, and second with the premise and hypothesis reversed. 
To better illustrate how the NLI is done, here is a quick example of the entailment classification: (ENT1, ENT2, length, 1) is one example in VPR where length means this is a comparison of length and 1 means ENT1 is longer than ENT2. 
Then we set the \neurocomps generated with this entity pair as the premise, and [‘ENT1 is longer than ENT2’, ‘ENT1 is shorter than ENT2’] as the hypothesis. 
Then entailment classification is run with each hypothesis and get respective entailment probability. 
The hypothesis with higher entailment probability \swabha{than what?} out of the two is chosen as the predicted entailment. 
If this predicted entailment complies with the original VPR example, we say our \neurocomps and the example in VPR entail each other, otherwise not. E.g., if ‘ENT1 is longer than ENT2’ has higher entailment probability, that means we our the example in the VPR is an entailment to our \neurocomps. 
This same process is done a second time but with premise being the example in VPR and hypothesis being our VPR (we just reverse the entities in our \neurocomps to get the counterfactual hypothesis). 
We count our \neurocomps as conformed / agreed by VPR if either of these two evaluations result in an entailment. 
Overall, we find that \neurocomps have a 66.7\% entailment rate in this study, which is consistent with our previous analysis on the accuracy of matching \neurocomps to the Elephant dataset (\S\ref{sec:downstream}).

\subsection{Comparative Reasoning QA}
\label{app-comp-reasoning-qa}
Here is an example of the QA transformation from comparative data:
\singlespacing
\begin{addmargin}[1em]{1em}
\small{
\textsl{\textbf{Comparative:} \newline Compared to graham crackers, kiwifruits can be considered healthier”,
\newline
\textbf{QA Format:} \newline Question: Which of the following can be considered healthier? \newline
A. graham crackers \newline
B. kiwifruits \newline
Answer: B
}}
\end{addmargin}
\singlespacing

\section{Terms and Licenses}
We use all of the datasets and models the way it is intended to be used if specified. For the ComparativeQA dataset we release, we intend its use for research only.

\begin{table*}[ht]
\centering
\small
\resizebox{\textwidth}{!}{%
\begin{tabular}{p{11.25cm}}
\toprule
\textbf{Few-shot Examples for Dimension Classification} \\
\midrule
Sentence: Compared to boats, trains are typically a lot slower. \\
Thought: Let's think step by step. we can analyze the words and their context to determine the appropriate class. In this case, the sentence compares the speed of boats and trains, implying that the class we should assign is "speed." \\ 
Action: Finish[speed] \\ \\
Sentence: Compared to cars, horses are often considered less practical. \\
Thought: Let's think step by step. "Less practical" specifically refers to the practicality of horses in comparison to cars. We can conclude that the appropriate class for this sentence is "practicality," as it discusses the comparative practicality of horses compared to cars. However, practicality is not one of the 4 classes (speed, length, mass, and others) we are supposed to choose from. Therefore, we should assign the class "others" to this sentence. \\
Action: Finish[others]\\\\
Sentence: Compared to chins, heads often have more muscle mass.\\
Thought: Let's think step by step. "More muscle mass" specifically refers to the muscle mass of heads in comparison to chins. Considering all these factors, we can conclude that the appropriate class for this sentence is "mass," as it discusses the comparative muscle mass of heads compared to chins.\\
Action: Finish[mass]\\\\
Sentence: Compared to fingers, arms are always longer.
Thought: Let's think step by step. "Longer" specifically refers to the length of arms in comparison to fingers. Considering all these factors, we can conclude that the appropriate class for this sentence is "length," as it discusses the comparative length of arms compared to fingers.\\
Action: Finish[length]\\\\
Sentence: Compared to dresses, purses can typically hold more goods.\\
Thought: Let's think step by step.  "smaller" specifically refers to the size or dimensions of purses in comparison to dresses. Considering all these factors, we can conclude that the appropriate class for this sentence is "size" or "dimensions," as it discusses the comparative size of purses compared to dresses.\\
Action: Finish[length]\\\\

\bottomrule
\end{tabular}
}
\caption{Few-shot examples used to classify NeuroComparatives into 4 dimensions: speed, length, mass and others.}
\label{tab:fewshot-classification-example}
\end{table*}

\begin{table*}[ht]
\centering
\small
\resizebox{\textwidth}{!}{%
\begin{tabular}{p{11.25cm}}
\toprule
\textbf{Comparative Adjectives} \\
\midrule
littler, denser, sweeter, dumber, itchier, rawer, skinnier, righter, bloodier, harder \\ 
wider, creepier, cheaper, sorrier, sillier, hairier, odder, worthier, idler, cooler \\ 
higher, sourer, softener, unhappier, sadder, stingier, hotter, busier, slimmer, narrower \\ 
subtler, sharper, shorter, sparser, lesser, needier, drier, greasier, pricklier, neater \\ 
lighter, cuter, shyer, sweatier, floppier, shadier, fitter, lazier, crazier, muddier \\ 
purer, sooner, nearer, fresher, further, louder, chubbier, whiter, crueler, thirstier \\ 
slighter, flakier, clumsier, greener, rougher, fatter, prettier, calmer, damper, politer \\ 
fiercer, messier, darker, poorer, lovelier, lower, handier, steeper, deadlier, jointer \\ 
greedier, cleverer, steadier, headier, blunter, blander, outer, younger, dirtier, wiser \\ 
direr, graver, greater, riper, milder, noisier, likelier, meaner, sneakier, unlikelier \\ 
tougher, upper, angrier, stronger, shinier, stricter, smoother, fuzzier, tenther, sorer \\ 
classier, fairer, gentler, brighter, trickier, grainier, looser, harsher, extremer, grander \\ 
juicier, guiltier, colder, ruder, tighter, sunnier, newer, stickier, wealthier, crankier \\ 
quicker, dustier, trendier, cleaner, rosier, richer, braver, prouder, shaggier, earlier \\ 
larger, lengthier, windier, fonder, sleepier, heartier, bluer, filthier, worser, taller \\ 
worse, spicier, heavier, quirkier, stockier, scarier, creamier, roomier, smarter, curlier \\ 
clearer, goofier, hardier, breezier, grosser, laster, firmer, mushier, quieter, chewier \\ 
plainer, jumpier, lonelier, madder, touchier, readier, smokier, mightier, bitterer, sexier \\ 
unhealthier, snowier, wilder, norther, closer, later, saner, crispier, flatter, nastier \\ 
deeper, briefer, finer, smaller, cozier, hungrier, curvier, tastier, bigger, happier \\ 
smellier, faster, simpler, easter, tinier, kinder, fainter, thinner, blacker, bolder \\ 
funnier, holier, weightier, poppier, sturdier, nobler, livelier, hipper, duller, fuller \\ 
slower, cloudier, rustier, rarer, wetter, coarser, better, leaner, firer, crunchier \\ 
gloomier, speedier, abler, riskier, warmer, blanker, soggier, nicer, keener, moister \\ 
shallower, yellower, stranger, weirder, stiffer, stupider, lousier, humbler, friendlier \\ 
stealthier, straighter, softer, bossier, icier, fancier, broader, uglier, nexter, loftier, naughtier \\ 
scarcer, worldlier, tanner, luckier, sincerer, bulkier, oilier, easier, warier, healthier \\ 
earthier, wobblier, less, more, choppier, swifter, longer, saltier, truer, weaker \\ 
older, fussier, steepler, fewer, safer, slimier, fattier, chillier, thicker, nimbler \\ 
\bottomrule
\end{tabular}
}
\caption{Full list of comparative adjectives (290 words).}
\label{tab:comparative-constraints-example}
\end{table*}

\begin{table*}[ht]
\centering
\small
\resizebox{\textwidth}{!}{
\begin{tabular}{p{11.25cm}}
\toprule
\textbf{Punctuation \& Nonsensical Characters (separated by tab)} \\
\midrule
\includegraphics[scale=0.43,trim={0 1.5cm 0 1cm},clip]{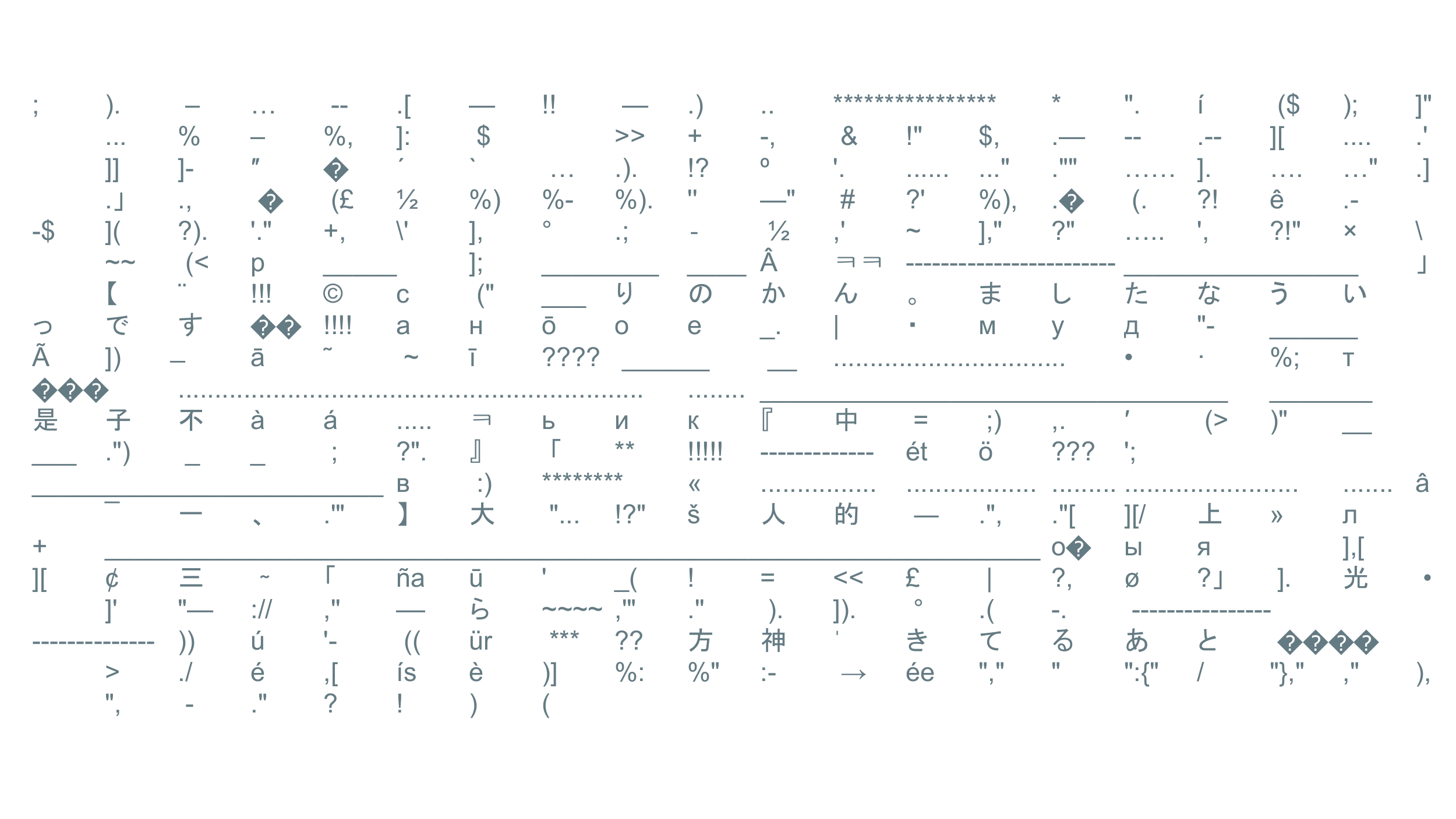} \\
\bottomrule
\textbf{Pronouns} \\
\midrule
I /  think /  you /  You /  He /  he /  he. /  They
 they /  they. /  she /  she. /  She /  my /  my. /  We /  we / \\
\bottomrule
\textbf{Discourse Connectives \& Relative Clause} \\
\midrule
 without /  without.
 between /  between. /  much /  much. /  either /  either. /  neither /  neither. /  and /  and. /  when
 when. /  while /  while. /  although /  although. /  am /  am. /  no /  no. /  nor /  nor.
 not /  not. /  as /  as. /  because /  because. /  since /  since. /  although /  although. /  finally
 finally. /  however /  however. /  therefore /  therefore. /  because /  because. /  consequently /  consequently. /  furthermore /  furthermore.
 nonetheless /  nonetheless. /  moreover /  moreover. /  alternatively /  alternatively. /  henceforward /  henceforward. /  nevertheless /  nevertheless. /  whereas
 whereas. /  meanwhile /  meanwhile. /  this /  this. /  there /  there. /  here /  here. /  same /  same.
 few /  few. /  similar /  similar. /  the following /  the following. /  by now /  by now. /  into /  into. /  than / than. / and \\
\bottomrule
\end{tabular}
}
\caption{Full list of negative constraint sets separated by "/".}
\label{tab:negative-constraints-full}
\end{table*}

\end{document}